\documentclass[pdflatex,sn-mathphys-num]{sn-jnl}

\usepackage{fix-cm}
\usepackage{mathrsfs}
\usepackage{graphicx}%
\usepackage{multirow}%
\usepackage{amsmath,amssymb,amsfonts}%
\usepackage{amsthm}%
\usepackage{mathrsfs}%
\usepackage[title]{appendix}%
\usepackage{xcolor}%
\usepackage{textcomp}%
\usepackage{manyfoot}%
\usepackage{booktabs}%
\usepackage{algorithm}%
\usepackage{algorithmicx}%
\usepackage{algpseudocode}%
\usepackage{listings}%
\usepackage{longtable}
\usepackage{caption}
\usepackage{hyperref}
\usepackage{cleveref}
\usepackage{url}
\usepackage{siunitx}
\usepackage{array}

\crefname{section}{Section}{Sections}
\Crefname{section}{Section}{Sections}

\begin{document}

\title[Article Title]{Toddlers' Active Gaze Behavior Supports Self-Supervised Object Learning}







\author[1,2]{\fnm{Zhengyang} \sur{Yu}}
\author[1,2]{\fnm{Arthur} \sur{Aubret}}
\author[1]{\fnm{Marcel C.} \sur{Raabe}}
\author[3]{\fnm{Jane} \sur{Yang}}
\author[3]{\fnm{Chen} \sur{Yu}}
\author[1]{\fnm{Jochen} \sur{Triesch}}

\affil[1]{\orgname{Frankfurt Institute for Advanced Studies}}
\affil[2]{\orgname{Xidian-FIAS International Joint Research Center}}
\affil[3]{\orgdiv{Department of Psychology}, \orgname{University of Texas at Austin}}

\email{\texttt{\{zhyu, aubret, raabe, triesch\}@fias.uni-frankfurt.de}}
\email{\texttt{\{jane.yang, chen.yu\}@austin.utexas.edu}}

\abstract{
Toddlers learn to recognize objects from different viewpoints with almost no supervision. During this learning, they execute frequent eye and head movements that shape their visual experience. It is presently unclear if and how these behaviors contribute to toddlers' emerging object recognition abilities. To answer this question, we here combine head-mounted eye tracking during dyadic play with unsupervised machine learning. We approximate toddlers’ central visual field experience by cropping image regions from a head-mounted camera centered on the current gaze location estimated via eye tracking. This visual stream feeds an unsupervised computational model of toddlers' learning, which constructs visual representations that slowly change over time. Our experiments demonstrate that toddlers’ gaze strategy supports the learning of invariant object representations. Our analysis also shows that the limited size of the central visual field where acuity is high is crucial for this. Overall, our work reveals how toddlers' gaze behavior may support their development of view-invariant object recognition.
}


\maketitle

\section{Introduction}\label{sec1}

Within their first year of life, children learn to map their visual inputs onto more abstract representations that support the recognition of objects observed from different viewpoints  \citep{kraebel2006three, ayzenberg2024development}. This early emergence of viewpoint-invariant recognition and the ease with which adults perform this skill hide the complexities of acquiring it. For example, retinal images vary drastically when objects are rotated in depth and even state-of-the-art machine learning methods make surprising recognition mistakes when faced with unusual viewpoints of objects \citep{dong2022viewfool, abbas2023progress, ruan2023towards}. 


One of the main theories for how children acquire invariant object recognition posits that their brains exploit a mechanism to construct visual representations that slowly change over time \citep{foldiak1991learning, li2008unsupervised, miyashita1988neuronal}. Toddlers abundantly manipulate (or walk around) objects while watching them, which gives them access to diverse views of a single object over a short period of time. By learning slowly changing representations, a toddler may be able to associate these different views, allowing them to form viewpoint-invariant representations of objects \citep{wiskott2002slow, schneider2021contrastive, aubret2022time}. 

When processing visual input, humans typically select only a limited portion of the central visual field of a few degrees of visual angle for detailed analysis \citep{quaia2024object, yu2015representation, zhaoping2024peripheral}. One reason for this is that receptor densities in the retina decline sharply towards the periphery \cite{jonas1992count, provis1985development}. As humans make on average around three saccades per second, the contents of the central visual field may be semantically unstable, i.e., the central visual field may contain frequent transitions between different objects. This might interfere with a learning mechanism based on slowness. However, toddlers may also actively move their gaze to stabilize the semantic content in their central visual field and support learning via a slowness objective. For example, a learning toddler may not move their gaze randomly within a scene, but watch an object they are manipulating for an extended period of time before saccading to a different object. Previous models of visual representation learning in infants and toddlers have neglected the importance of eye gaze for learning \cite{orhan2020self, orhan2024learning, orhan2024self}.

Here, we investigate whether toddlers' gaze behaviors may support the learning of view-invariant object representations. To this end, we leverage a dataset of head-camera video recordings and eye gaze tracking from toddlers and adults during play sessions \citep{bambach2018toddler}. To simulate participants' central visual field experience, we extract image patches centered on tracked gaze locations. This data feeds a computational model of a toddler's visual representation learning, which constructs representations that slowly change over time \citep{aubret2022time,schneider2021contrastive}. Our results show that toddlers' gaze strategies boost visual learning in comparison to several baselines. Furthermore, we demonstrate that restricting learning to input from the central visual field improves the emerging object representations. Finally, we show that the visual input from toddlers permits learning better representations than that from adults. This finding may be explained by toddlers looking longer at individual objects while manipulating them. Taken together, our study reveals how gaze behaviors and the principle of temporal slowness may jointly underpin the development of invariant object recognition abilities in humans.

\begin{figure}[ht]
  \centering
  \includegraphics[width=1\textwidth]{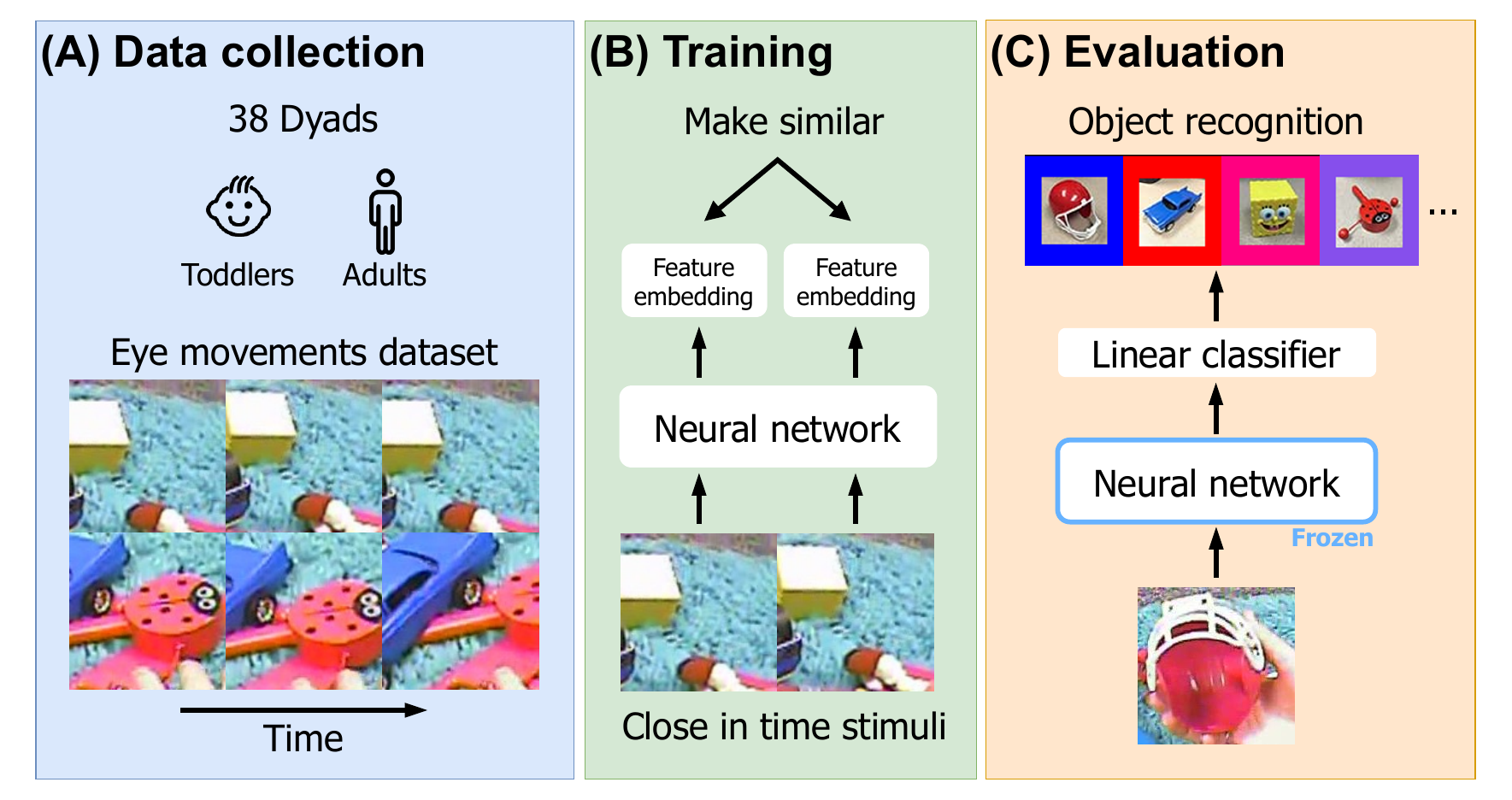}
  \caption{Overview of the experimental framework. {\bf A.} Eye-tracking data were collected from 38 toddlers and their caregivers. The resulting eye movement sequences formed a temporal dataset of visual observations. {\bf B.} We trained a computational model of biological visual learning to make the representations of temporally adjacent frames more similar. {\bf C.} To assess the quality of the learned visual representations, we train a linear classifier on top of the frozen trained neural network to perform object recognition.}
  \label{fig:main}
\end{figure}

\section{Method}
\figureautorefname~\ref{fig:main} provides an overview of the dataset, our computational model of toddlers visual learning, and the evaluation procedure used in this study. In the following sections, we briefly describe each component. More details are supplied in \appendixautorefname~\ref{app:method}.

\subsection{Datasets}
\label{sec:datasets}

We build on a dataset containing head-camera videos and eye-tracking data recorded from 38 dyads of toddlers and caregivers. Each video shows a dyad that plays with the same 24 toys for 15 minutes on average. To simulate the central visual field experience of a participant based on their eye gaze, we extract image patches (corresponding to $\SI{14}{\degree} \times \SI{14}{\degree}$ of visual angle) from video frames of the scene camera mounted on the participant's head, which are centered on the recorded gaze position. \figureautorefname~\ref{fig:toddler_fixation}(A) shows an example of a sequence of image patches extracted around subsequent gaze locations \citep{bambach2018toddler}. We compare learning based on the visual input streams produced by these measured gaze behaviors (``Toddlers'/Adults' eye movements'') of dyads against learning based on two simulated alternative gaze strategies. The first assumes that the camera-wearer samples gaze locations uniformly at random within the field of view of the scene camera (``Random eye movements''). The second ignores eye movements (as in previous works) and instead assumes that the gaze location always remains in the center of the participant's field of view (``No eye movements''). 
In addition, we consider two idealized ``oracles'' that simulate a (biologically implausible) learner who fixates only on the objects and does not learn during transitions between objects. I.e., this hypothetical learner already has perfect knowledge of when it is looking at an object of interest. We consider this strategy with natural backgrounds (``Object fixation'') and with blank backgrounds (``Blank background''). We show examples of visual sequences resulting from these strategies in \figureautorefname~\ref{fig:toddler_fixation}.

\subsection{Computational model}
\label{sec:tt}
To model the learning process of humans, we train deep neural networks with two bio-inspired self-supervised learning models,  namely SimCLR-TT and BYOL-TT \cite{schneider2021contrastive}. Both models learn visual representations that associate close-in-time visual inputs. They also include a hyper-parameter $\Delta T$ (measured in seconds) that quantifies how slowly representations should change. We use a ResNet18 as our default neural network architecture and provide additional results with a ResNet50 in \appendixautorefname~\ref{appendix:encoder}. Networks are trained ``from scratch,'' i.e., they are not already pretrained for object recognition or any other task. \figureautorefname~\ref{fig:toddler_fixation}(B) illustrates the learning process.

\subsection{Evaluation} 
We train the models with video frames recorded from 30 randomly chosen dyads and keep the other 8 for testing. We also consider training on the recording of single participants. We assess the quality of the learned representations by training a linear classifier on top of the learned representation (right after the average pooling layer) in a supervised fashion \citep{chen2020simple}. Since our model of human visual representation learning does not use labeled images, we always train the linear classifier on the train split of the Objects Fixation dataset, which was manually labeled, and evaluate the object recognition accuracy on the test split of the Objects Fixation dataset. Statistical analyses were performed to assess the results. Unless otherwise specified, all significance tests were performed using two-tailed independent two-sample t-tests. Correlation analyses were conducted using Pearson’s correlation coefficient.

\section{Results}

\begin{figure}[h]
  \centering
  \includegraphics[width=\textwidth]{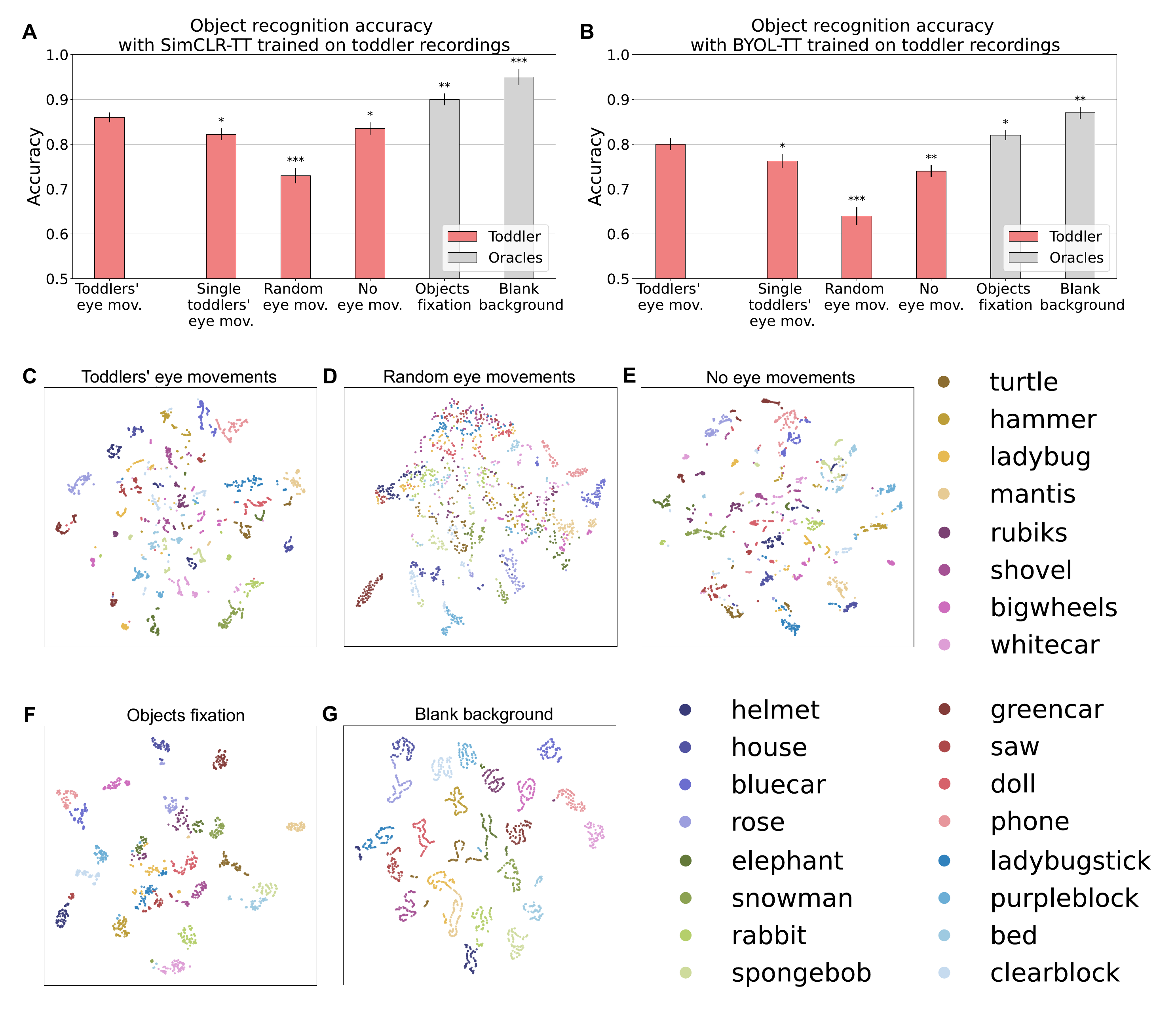}
  \caption{Toddlers' gaze behavior supports the learning of invariant object representations. {\bf (A)} and {\bf (B)} show classification results of the linear classifier trained on top of representations learned with SimCLR-TT and BYOL-TT, respectively. Error bars represent the standard deviation over three random seeds. When training on data from just one toddler (``Single toddlers' eye mov.''), we show the average accuracy and the standard deviation over the 38 toddlers. Results of statistical tests comparing toddlers' eye movements to other conditions are indicated as follows: ``***'': $\text{p-value} < 0.001$, ``**'': $\text{p-value} < 0.01$, and ``*'': $\text{p-value} < 0.05$. {\bf (C)-(G)} t-SNE visualizations of learned object representations for SimCLR-TT trained on different datasets. Each point represents an image sample, color-coded by object identity (24 in total).} 
  \label{fig:acc_pretrained}
\end{figure}

\subsection{Toddlers' central visual field experience supports the learning of invariant object representations via time-based self-supervised learning}
\label{sec:bio}
To test whether toddlers' gaze behavior supports the learning of strong object representations, we compare the representations learned by the two unsupervised learning models based on slowness (SimCLR-TT and BYOL-TT) when trained on the different datasets introduced in \cref{sec:datasets}. \figureautorefname~\ref{fig:acc_pretrained}A and B show that models trained with the Toddlers' Eye movements dataset outperform those trained with the Random Eye movements dataset or the No Eye movements dataset (t-tests, $ p < 0.05 $ in all cases). This suggests that toddlers' gaze behavior supports the learning of view-invariant object representations. A similar result is observed for adults (\appendixautorefname~\ref{appendix:adults}). 

Next, we study the impact of toddlers' gaze behaviors on the learnt visual representations at a qualitative level. For each model trained using different eye movement strategies, we extract their representations of images in the Objects Fixation dataset and project these representations into a 2-dimensional embedding space using t-SNE \citep{van2008visualizing}. This allows us to visualize the (dis)similarity between representations of views of different objects (\figureautorefname~\ref{fig:acc_pretrained}(C)-(G)).
The model trained on the Toddlers’ Eye movements dataset shows well-separated clusters, indicating strong and discriminative object representations. In contrast, models trained with the Random or No Eye movements datasets produce less distinct clusters, reflecting weaker object learning. 
In contrast, the two biologically implausible ``oracle'' methods that train with the Objects Fixation dataset (F) or the Blank Background dataset (G) exhibit improved clustering.


We wondered whether the visual experience of only \textbf{a single} toddler during a play session suffices to build good visual representations. To investigate this question, we train SimCLR-TT on the individual recordings of each toddler and compute the average of linear accuracies. We train the neural network using all fixation data from a single toddler, followed by training and testing the linear classifier with the Objects fixation data from the same and different toddlers. We control the training set to comprise 75\% of the total data, ensuring that the test set does not overlap with the training set. \figureautorefname~\ref{fig:acc_pretrained}A and B show that the central visual experience of one toddler leads to representations almost as good as those resulting from the central visual experience of all toddlers.

\subsection{Constraining input to the central visual field improves learning}
\label{sec:fovea}
Previous computational studies of learning from infants' first person visual experience have neglected the importance of the constrained size of the central visual field for learning visual representations \citep{orhan2020self,sheybani2024curriculum}. Here, we assess whether our simulated central visual field experience leads to better/worse object representations than learning from a wide field of view. We vary the size of the image portion extracted to simulate central vision. In \tableautorefname~\ref{tab:size}, we observe for both toddlers and adults that an image size of $128\times128$ (corresponding to $\SI{14}{\degree} \times \SI{14}{\degree}$ of visual angle) produces the best recognition accuracy for all gaze strategies. Importantly, results for toddlers' and adults' eye movements at an image size of $128\times128$ pixels present an accuracy boost of $8\%$ compared to the no eye movements condition for an image size of $480\times480$ pixels, which simulates head-camera recordings without eye-tracking information. We conclude that accounting for the constrained size of the central visual field is crucial for learning powerful object representations. We speculate that this boost results from the property of an $128\times128$ gaze-centered crop to frequently capture the complete structure of an object while minimizing irrelevant background information --- at least in the used dataset.

\begin{table}[h]
\caption{Linear object recognition accuracy for different cropping sizes. Best results in each row are highlighted in bold. Underlined results represent simulations that do not utilize actual gaze fixations and consider only the egocentric visual experience.}
\vspace{0.5em}
\tabcolsep=0.2cm
\centering
\renewcommand\arraystretch{1.5}
\begin{tabular}{cccccc}
\toprule
&  & $64\times64$  & $128\times128$  & $240\times240$ & $480\times480$ \\ \hline
\multirow{2}{*}{Humans eye mov.}
& Toddler 
& $0.831\pm{0.015}$
& $\textbf{0.863}\pm{\textbf{0.011}}$
& $0.828\pm{0.014}$ 
& $0.805\pm{0.018}$ \\                               
& Adult  
& $0.826\pm{0.013}$ 
& $\textbf{0.851}\pm{\textbf{0.028}}$  
& $0.816\pm{0.013}$ 
& $0.791\pm{0.019}$ \\\hline
\multirow{2}{*}{Random eye mov.}   
& Toddler 
& $0.701\pm{0.011}$ 
& $\textbf{0.736}\pm{\textbf{0.017}}$ 
& $0.694\pm{0.025}$ 
& $0.589\pm{0.036}$ \\
& Adult  
& $0.716\pm{0.021}$ 
& $\textbf{0.742}\pm{\textbf{0.022}}$
& $0.685\pm{0.023}$ 
& $0.576\pm{0.019}$ \\\hline
\multirow{2}{*}{No eye mov.} 
& Toddler 
& $0.822\pm{0.016}$ 
& $\textbf{0.838}\pm{\textbf{0.010}}$ 
& $0.815\pm{0.018}$ 
& $\underline{0.784}\pm{\underline{0.022}}$ \\
& Adult  
& $0.818\pm{0.012}$ 
& $\textbf{0.829}\pm{\textbf{0.009}}$ 
& $0.807\pm{0.014}$ 
& $\underline{0.763}\pm{\underline{0.017}}$ \\ 
\bottomrule
\end{tabular}
\label{tab:size}
\end{table}

\subsection{Toddlers' gaze behavior favors stronger emphasis on slowness}
\label{sec:mod}
\begin{figure}[ht]
  \centering
  \includegraphics[width=1\textwidth]{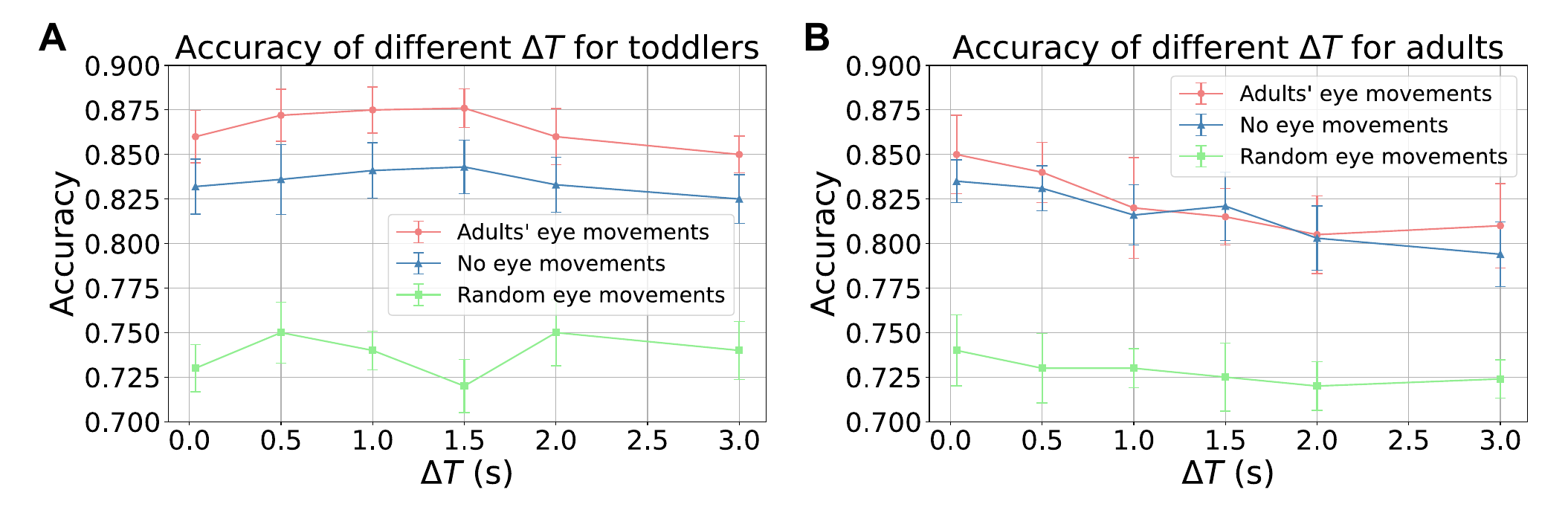}
  \caption{The impact of different $\Delta{T}$ on recognition accuracy for {\bf (A)} toddlers and {\bf (B)} adults. Error bars represent the standard deviation over three random seeds.}
  \label{fig:acc_skip}
\end{figure}

Semantic aspects of the visual experience have been shown to vary more slowly for toddlers than for adults \citep{sheybani2023slow}. Our learning model includes a hyper-parameter $\Delta T$ (measured in seconds) that quantifies how slowly representations should change. Concretely, it specifies the (maximum) time interval between two positive input pairs (whose representations will be made more similar) in the learning algorithms (see \appendixautorefname~\ref{appendix:learning} for details). Interestingly, previous work has shown that increasing $\Delta T$ can improve the quality of object representations if visual inputs are sufficiently stable over time \citep{aubret2022time,schneider2021contrastive}. Thus, we wondered how changing $\Delta T$ may affect results and whether it may amplify any differences in the quality of representations learned from toddlers' vs.\ adults' first person visual experience. To test this, we varied $\Delta{T}$ from $\frac{1}{30}$ to 3.0 seconds (\figureautorefname~\ref{fig:acc_skip}A). The models trained with the Toddlers' Eye movements dataset (red) achieve the highest recognition accuracy for an intermediate value of $\Delta{T}=1.5$s. In contrast, \figureautorefname~\ref{fig:acc_skip}B shows that, for models trained with the Adults' Eye movements dataset, increasing $\Delta T$ only decreases the quality of the learned object representations. The results are consistent for both human eye movements (red) and the no eye movements datasets (blue). We conclude that toddlers' gaze behavior favors a stronger emphasis on slowness (greater $\Delta T$) than that of adults.

\subsection{Toddlers' long object inspections relative to adults facilitate learning}
\label{sec:metric}


So far, we have shown that the gaze behaviors of humans support object learning  via an unsupervised slowness objective and that toddlers gaze behavior favors a stronger emphasis on slowness (greater $\Delta{T}$). We wondered what differences between toddlers' and adults's eye movements may contribute to this discrepancy. We analyzed four metrics that characterize the temporal sequence of images: the average fixation duration, the average duration of bouts of looking at the same object when not holding the object, the average duration of looking at an object when holding it, and the cumulative duration of object looking in an entire recording session.
For these analyses, we leverage manually labeled timestamps (by \citep{bambach2018toddler}) about when toddlers and adults look at/hold an object. See \appendixautorefname~\ref{appendix:cal_saccade} for details on saccade detection and calculation of fixation durations.

We successfully extracted the data from 28 out of 38 toddlers and conducted all subsequent experiments using these 28 toddlers. The remaining participants are excluded from this analysis due to the lack of data on fixation durations. \tableautorefname~\ref{tab:toddler} in \appendixautorefname~\ref{appendix:data} presents the details of the 28 included toddlers.

\begin{figure}
  \centering
  \includegraphics[width=1\textwidth]{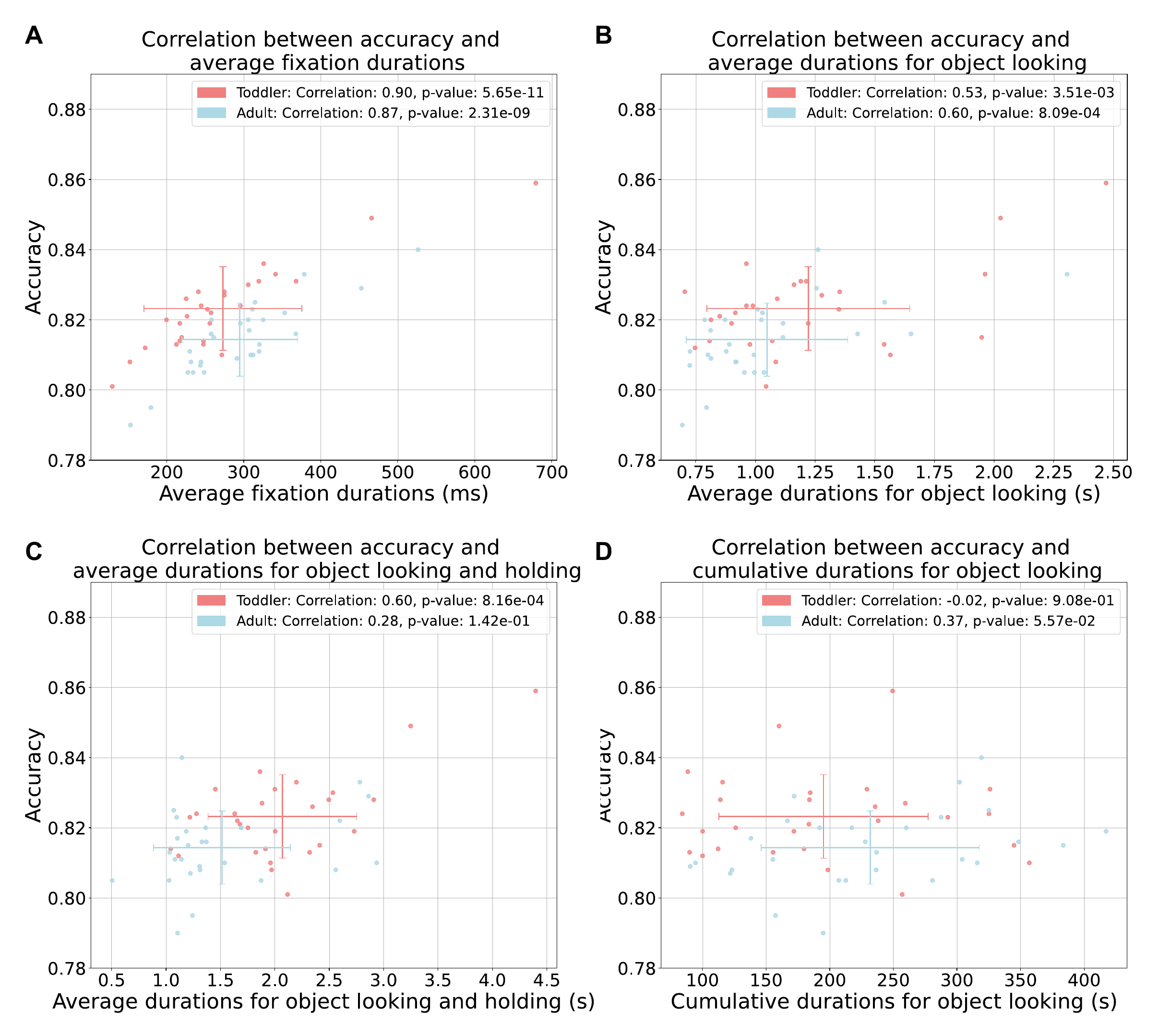}
  \caption{Correlation analysis between the recognition accuracy and {\bf (A)} the average fixation duration; {\bf (B)} the average duration of object looking while not holding the object; {\bf (C)} the average duration of object looking while holding the object and {\bf (D)} the cumulative duration of object looking. Models were all trained on the individual Toddlers' and Adults' Eye movements dataset. In each figure, the crosshairs represent the mean and standard deviation of the data values over the two axes. The legends show the Pearson correlation coefficients and their p-values.}
  \label{fig:pearson}
\end{figure}

\begin{figure}
  \centering
  \includegraphics[width=1\textwidth]{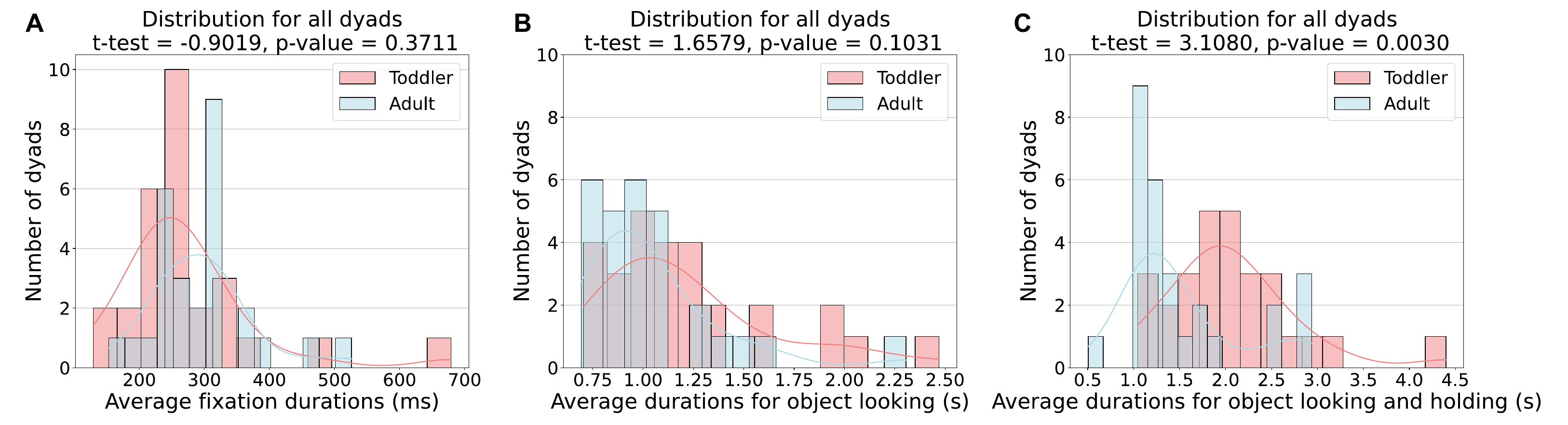}
  \caption{Comparison of {\bf (A)} average fixation duration, {\bf (B)} average duration of object looking while not holding the object, and {\bf (C)} average duration of object looking while holding the object for toddlers and adults. Each panel includes the frequency distribution for the given metric, along with a density curve.}
  \label{fig:stat_more}
\end{figure}

In \figureautorefname~\ref{fig:pearson}, we observe that object recognition accuracy is highly correlated with fixation durations, durations of object looking and duration of object looking while holding the object, but only weakly correlated with the cumulative duration of object looking. This indicates that long fixation bouts on the same object are important in explaining the relative quality of visual representations trained on Toddlers' and Adults' Eye movements datasets.

\figureautorefname~\ref{fig:pearson} also shows that on average toddlers' visual experience permits learning better representations than that of adults (t-tests, $ p=0.0053 $) with the given subset of dyads. To investigate which metric plays a crucial role in these differences. \figureautorefname~\ref{fig:stat_more} compares the distributions of average fixation durations for toddlers and adults. The t-test statistics and p-values are given in the titles. We observe that toddlers look longer at the object that they are holding (t-test, $p=0.003$). Other metrics do not exhibit statistically significant differences between adults and toddlers. We conclude that, compared to adults, toddlers’ longer periods of object inspection while manipulating objects allow for learning better viewpoint-invariant object representations.

\section{Conclusion}
The mechanisms permitting infants and toddlers to acquire sophisticated object recognition abilities from very limited first person visual experience are still poorly understood.
Here, we investigated whether neuro-biologically motivated models of visual learning can take advantage of toddlers' gaze behavior to develop robust object representations. We extracted toddlers' gaze locations from egocentric video recordings with head-mounted eye-tracking during play sessions. Then, we trained self-supervised deep learning models that drive visual representations to change slowly with the extracted first person experience of toddlers. Our findings indicate that toddlers' gaze strategies permit the learning of representations that support view-invariant object instance recognition within a single play session of 12 minutes. Interestingly, models trained on adults' visual experience performed significantly worse. Our analysis showed that focusing learning on inputs from the central visual field is beneficial for learning viewpoint-invariant object representations and that toddlers' gaze behavior favors a stronger emphasis on slowness compared to that of adults. This is consistent with toddlers looking longer at objects while holding them. During these relatively long holding periods, toddlers often turn and move the object, giving them access to sequences of different views of an object over a short period of time.

From a developmental perspective, our work provides strong evidence that the development of viewpoint-invariant representations can originate from a slowness learning objective, a mechanism supported by neuroscientific studies \citep{li2008unsupervised, miyashita1988neuronal}. Our results also suggest that toddlers curate their gaze behavior to enhance their learning of visual representations.
From a machine learning perspective, we show that combining head-mounted eye-tracking video data with time-based self-supervised learning supports the emergence of viewpoint-invariant object recognition. Our work therefore marks a significant step toward learning strong visual representations without handcrafted image datasets (e.g., \citep{aubret2022time}).

Our work has several limitations.
We analyzed gaze behavior of toddlers with a minimum age of 12.3 months, meaning they had substantial visual learning experience before the experiment, while our computational models learned from scratch. Expanding to younger toddlers and a more diverse visual diet and distinct visual exploration patterns, could offer further insights into early visual representation development. Studying how babies under one year engage with objects may reveal new aspects of gaze behavior that contribute to their visual learning \citep{maurer2017critical, sheybani2024curriculum}. Such attempts should also be guided by knowledge of infants' developing contrast sensitivity. Moreover, refining our approach to utilize both central and peripheral vision for learning visual representations could provide a more accurate simulation of the development of object vs.\ scene representations in different brain areas. Finally, a more complete model of visual learning in infants and toddlers needs to also capture the computational mechanisms driving their gaze shifts, whose relevance for visual representation learning we have demonstrated here. Understanding these mechanisms will be an important next step in unraveling the mechanisms underlying the early development of human visual perception \citep{wang2021use}.


\backmatter
\section*{Declarations}
\bmhead{Acknowledgements}
ZY thanks the Xidian-FIAS International Joint Research Center for funding. This research was supported by “The Adaptive Mind” funded by the Excellence Program of the Hessian Ministry of Higher Education, Research, Science and the Arts, Germany, and by the Deutsche Forschungsgemeinschaft (DFG project “Abstract REpresentations in Neural Architectures (ARENA)”). This work was funded by NIH R01HD074601 and R01HD093792 to CY. JT was supported by the Johanna Quandt foundation. We thank the members of the Developing Intelligence Lab at UT Austin and participating families for their contributions. We gratefully acknowledge support from Goethe University (NHR Center NHR@SW) for providing some of the computing and data-processing resources needed for this work.

\bmhead{Data availability}
The data supporting our analyses were obtained from the private dataset described in \citep{bambach2018toddler}. Anyone interested in accessing the data may contact the Developing Intelligence Lab at The University of Texas at Austin (chenyulab@austin.utexas.edu) for further assistance.

\bmhead{Code availability}
The Python code used for data processing (including different cropping strategies), analysis, and results in this study is publicly available on GitHub (https://github.com/jestland/infantVision).

\bibliography{sn-bibliography}

\clearpage









\begin{appendices}

\section{Method details}
\label{app:method}
\subsection{Datasets}\label{app:datasets}

\begin{figure}
  \centering
  \includegraphics[width=1\textwidth]{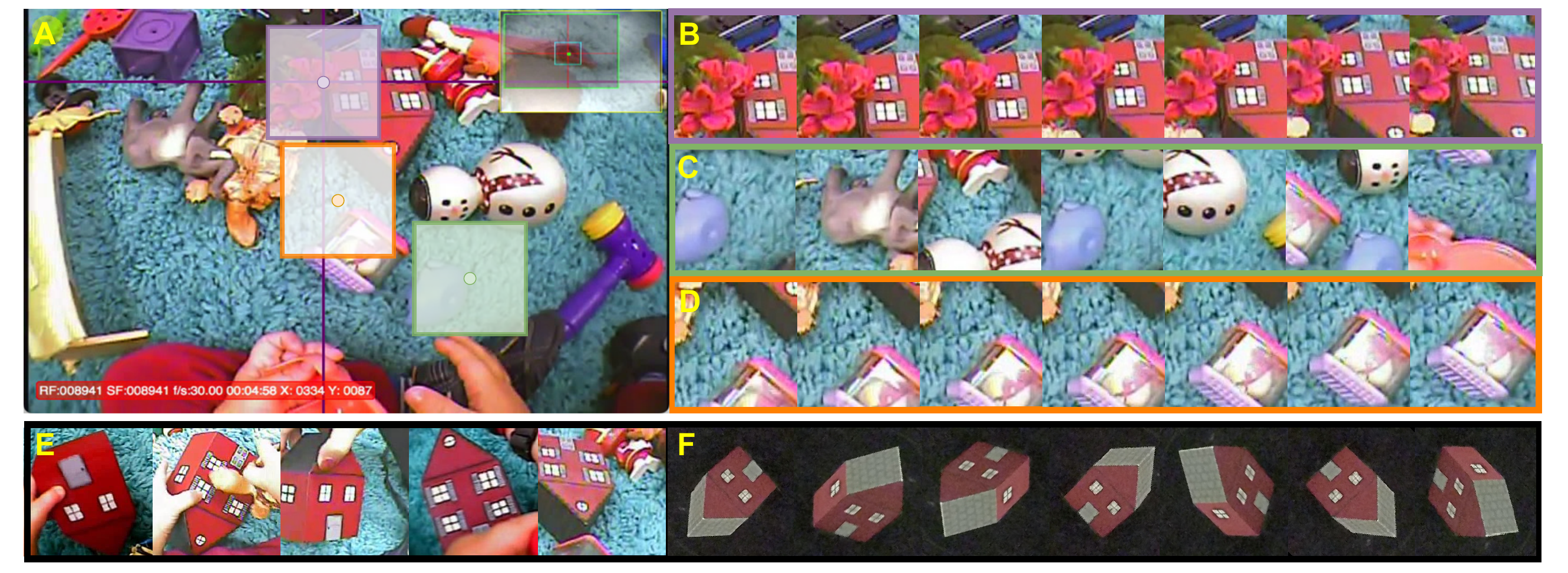}
  \caption{Examples of visual sequences for each of our datasets. {\bf A.} Raw frame from an egocentric video with the locations of our different croppings. Purple, orange, and green boxes represent the real-time location of human eye movements, no eye movements, and random eye movements, respectively. The cross indicates the gaze location given by the eye-tracker. {\bf B-F.} Example sequences for {\bf B-} the Toddlers' Eye movements dataset; {\bf C-} the Random Eye movements dataset; {\bf D-} the No Eye movements dataset; {\bf E} the Objects Fixation dataset and {\bf F-} the Blank Background dataset. Note that datasets {\bf E-F} have been manually curated to only contain views of the target objects. This kind of oracle knowledge is not available to a naive learner.}
  \label{fig:toddler_fixation}
\end{figure}

The used dataset \citep{bambach2018toddler} contains head-camera videos recorded at 30 frames per second (FPS) and eye-tracking data for 38 dyads of toddlers and caregivers. All dyads play with the same set of 24 toys for 12 minutes on average. The toddlers' ages range from 12.3 to 24.3 months. For 30 dyads, a head-camera resolution of 640 $\times$ 480 pixels was used, while four dyads were recorded at 720 $\times$ 480 pixels and the remaining four at 320 $\times$ 240 pixels. The horizontal field of view covers 72 degrees. \figureautorefname~\ref{fig:toddler_fixation}A shows an example video frame with the gaze location \citep{bambach2018toddler}. In the following, we explain how we simulate different gaze strategies by deriving several datasets from these videos and gaze positions. Additionally, we include the anonymized information of all toddlers who participated in the study in \appendixautorefname~\ref{appendix:data}.

\textbf{Toddlers' Eye movements dataset.} This dataset aims to simulate the central visual experience of toddlers. We cut out an image patch centered on the current gaze point.
For the cut-out's size, we choose 128 $\times$ 128 pixels as the default, which corresponds to $\SI{14}{\degree} \times \SI{14}{\degree}$ of visual angle. A typical temporal sequence of this dataset is illustrated in \figureautorefname~\ref{fig:toddler_fixation}B. If the gaze fixation point is too close to the image border, the crop boundaries may extend beyond the image, making it impossible to extract a patch of the desired size. In this case, we shift the gaze fixation point from the problematic border orthogonally by the minimum number of pixels. This ensures that the cropping operation outputs an image with the correct size. Note that the cropped area always contains the actual gaze fixation point. This dataset contains 559,522 training images. This number is consistent across all eye movement datasets (see below).

\textbf{Adults' Eye movements dataset.} We want to investigate the differences between gaze fixation in adults and toddlers and the consequences of these differences on learned representations. Thus, we also extract image patches around adults' gaze fixation points following the procedure of the Toddlers' Eye movements dataset. \appendixautorefname~\ref{appendix:gaze} illustrates the gaze distributions of toddlers and adults.

\textbf{Random Eye movements dataset.} As a comparison dataset, we simulate a completely random gaze strategy. We crop each frame around a location that is sampled uniformly at random. Unlike the Toddlers'/Adults' Eye movements datasets, this dataset shows little spatio-temporal correlation structure, and the cropped images are unlikely to contain well-centered objects. \figureautorefname~\ref{fig:toddler_fixation}D provides example frames from the Random Eye movements dataset.

\textbf{No Eye movements dataset.} We also propose a stronger comparison dataset that considers a human moving their head but not their eyes. This is an important comparison because it distills the effect of eye movements. One possibility for building such a dataset could be to always crop the center of the frames. However, we noticed that the head-camera was often misaligned with respect to the stationary position of the eyes, resulting in a mismatch between the center of the frames and the center of the camera wearer's field of view (cf.\ \appendixautorefname~\ref{appendix:gaze}). Thus, we rather use the centroid of the gaze fixation points (one for each participant video). To compute these centroids, we gather all gaze fixation points of a subject and calculate the mean of their horizontal and vertical coordinates. Note that, despite the centroid positions being fixed, the continuous movements of the head change the visible portions of the scene. Nevertheless, compared to the Random Eye movements dataset, this set contains image patch sequences that are relatively stable over time. \figureautorefname~\ref{fig:toddler_fixation}D presents a temporal sequence of the No Eye movements dataset.

\bigskip
We also consider two ``oracle'' datasets that were constructed using the ground truth about an object's identity/location. Models trained on this dataset aim to upper-bound our model's performance.

\textbf{Objects Fixation dataset.} This dataset was collected from the same video frames used in the Toddlers' Eye movements dataset. Images were manually filtered so that toddlers looked at one of the target objects. From these frames, crops with a 30-degree field of view around the gaze location were extracted, containing the target object while minimizing background interference \citep{bambach2018toddler,tsutsui2021reverse}. This dataset contains 271,754 images. \figureautorefname~\ref{fig:toddler_fixation}E displays examples of images. 

\textbf{Blank Background dataset.} This dataset contains 1,536 images, capturing each object from 128 possible viewpoints with various angles and distances. All toy models in this dataset originate from the Object Fixation dataset. Each image displays a complete object against a black background, ensuring visual isolation from external distractions. \figureautorefname~\ref{fig:toddler_fixation}E shows an example toy from different viewpoints.

\subsection{Learning models}
\label{appendix:learning}
To model the learning process of humans, we learn visual representations with self-supervised models that construct similar representations for close-in-time visual input.

\begin{figure}
  \centering
  \includegraphics[width=1\textwidth]{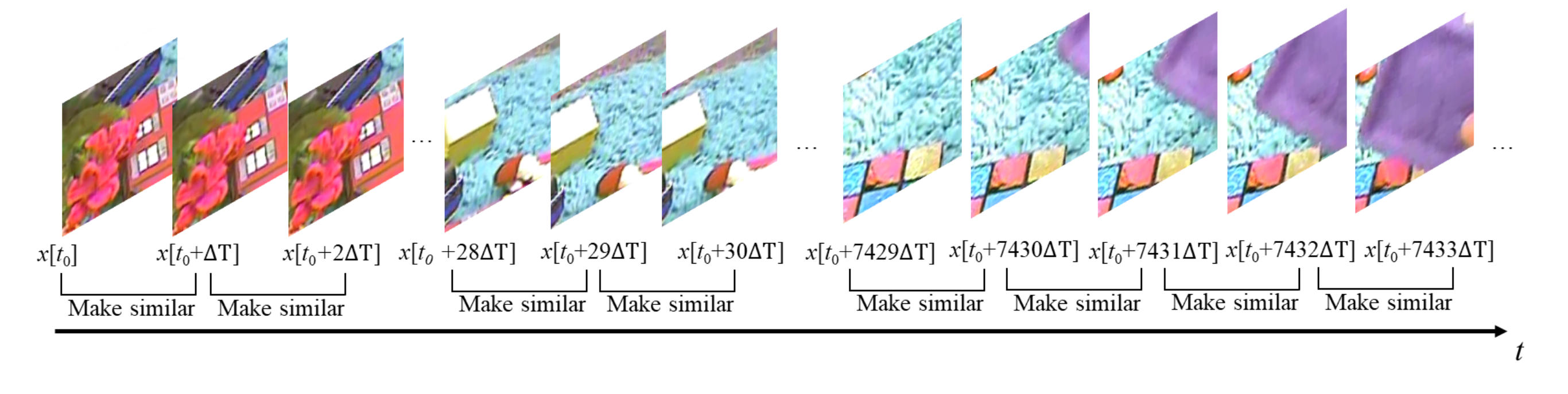}
  \caption{Illustration of SimCLR-TT on the Toddlers' Eye movements dataset. By default, the time interval $\Delta{T}=\frac{1}{30}$s corresponds to the inverse of the camera's frame rate, but it can be increased to an integer multiple of this value.}
  \label{fig:sample}
\end{figure}

\textbf{SimCLR-TT.} This algorithm is based on the state-of-the-art SimCLR method \citep{chen2020simple}. SimCLR-TT samples an image $x_t$ at time $t$ and a temporally close image $x_{t+\Delta T}$ and computes their respective embeddings $z_t$, $z_{t+\Delta T}$ with a deep neural network \citep{schneider2021contrastive} (e.g.\ a ResNet). Unless stated otherwise, we set $\Delta{T}$ to the inverse of the camera's frame rate, i.e., $\Delta{T}=\frac{1}{30}$ seconds. 
\textcolor{black}{In \cref{sec:mod} we show additional results varying $\Delta T$.}
Then, SimCLR-TT minimizes
\begin{equation}
\color{black}
\mathcal{L}\left(z_t,z_{t+\Delta T}\right)=-\log \frac{\exp \left(\operatorname{sim}\left(z_t, z_{t+\Delta T}\right) / \tau\right)}{\sum_{z_k \in \mathcal{B}, k \neq t}^{}\left[\exp \left(\operatorname{sim}\left(z_t, z_{k}\right) / \tau\right)\right]},
\end{equation}
where $\mathcal{B}$ is a minibatch, ${\rm sim}(\cdot)$ is the cosine similarity and $\tau$ is the temperature hyper-parameter. Here $k \neq t$ but $k = t+\Delta T$ is possible. Thus, SimCLR-TT maximizes the similarity between temporally close representations (numerator) while keeping all representations dissimilar from each other (denominator). \figureautorefname~\ref{fig:sample} illustrates the learning process of SimCLR-TT.

\textbf{BYOL-TT.} To evaluate whether our conclusions also hold for different methods that learns with temporal slowness, we perform the same experiments with BYOL-TT. Similarly to SimCLR-TT, BYOL-TT was originally considered to be used for contrastive learning through time \citep{schneider2021contrastive}. Its loss function is defined as
\begin{equation}
    \mathcal{L}_{\theta_{t}, \xi_{t+\Delta{T}}}=2-2 \cdot \operatorname{sim}\left(\mathrm{q}_{\theta_{t}}(\mathrm{z}_{\theta_{t}}\right),
    \mathrm{z}_{\xi_{t+\Delta{T}}}),
\end{equation}

where \( q_{\theta_{t}}(z_{\theta_{t}}) \) is the prediction of the online network for one frame and \( z_{\xi_{t+\Delta{T}}} \) represents outputs from the target network. The weights of the online network are denoted by \( \theta \), and \( \xi \) represents the weights of the target network. Again, we use the cosine similarity as the similarity function. \appendixautorefname~\ref{appendix:gaze} provides details on the training and evaluation procedures.

\subsection{Extraction of saccades and fixations} The analyses in  \cref{sec:metric} include extracting fixation bouts. This requires detecting saccades, as they mark possible beginnings and ends of the fixation bouts. To detect saccades in gaze movements, we apply a velocity threshold-based method similar to \citep{raabe2023saccade}. Consecutive gaze points that exceed a threshold $T_1$ are identified as a single saccade. To account for artifacts caused by low frame rates, a second threshold $T_2$, along with an angular criterion $\theta$, allows the inclusion of the two data points adjacent to the saccade initially detected. Any data points not classified as saccades are considered fixations. For this study, we choose $T_1=\SI{25}{\degree \per \second}$, $T_2=\SI{10}{\degree \per \second}$ and $\theta=\SI{45}{\degree}$. 
\label{appendix:cal_saccade}

\subsection{Neural network training details} 
We ran three random seeds for all experiments. For each seed, we split the 38 available dyads into 30 train dyads and 8 test dyads. We train the models on train dyads for 100 epochs with a ResNet18, the AdamW optimizer and set the initial learning rate and weight decay to $10^{-2}$ and $10^{-4}$, respectively. We set the SimCLR temperature to $0.08$ and the batch size to $256$. \appendixautorefname~\ref{appendix:hp} presents the results under various settings of hyper-parameters. We conduct all experiments on an Nvidia GeForce RTX 3090 GPU with 24 GB memory.

\section{Complementary analyses}\label{appendix:gaze}


\subsection{Gaze location distribution} 
\begin{figure}[ht]
  \centering
  \includegraphics[width=1\textwidth]{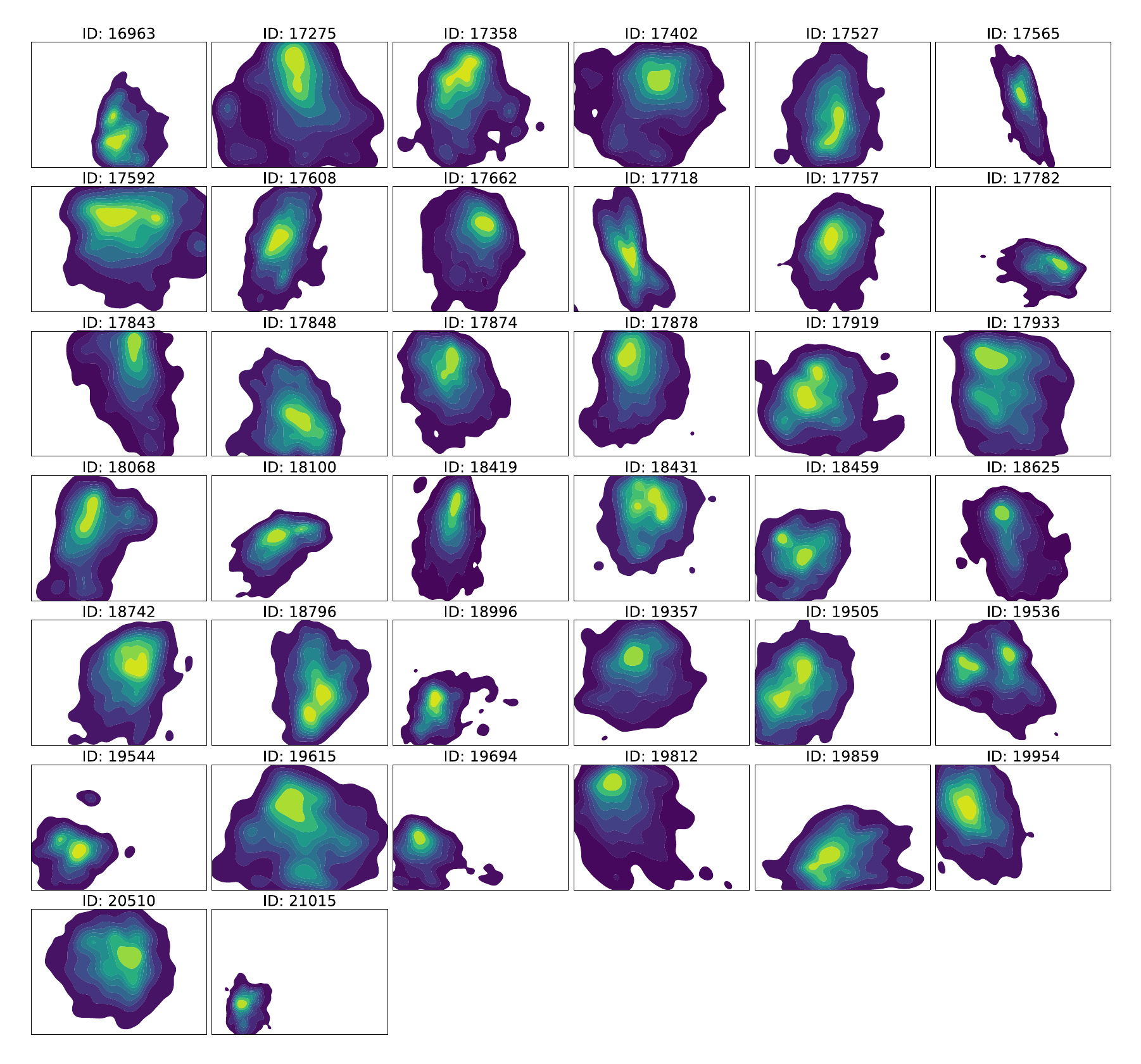}
  \caption{Gaze distribution of individual toddlers during dyadic play. Each subplot shows a toddler's 2D gaze distribution (kernel density estimate) across a play session. Warmer colors indicate higher gaze density.}
  \label{fig:gaze_toddlers}
\end{figure}

\begin{figure}[ht]
  \centering
  \includegraphics[width=1\textwidth]{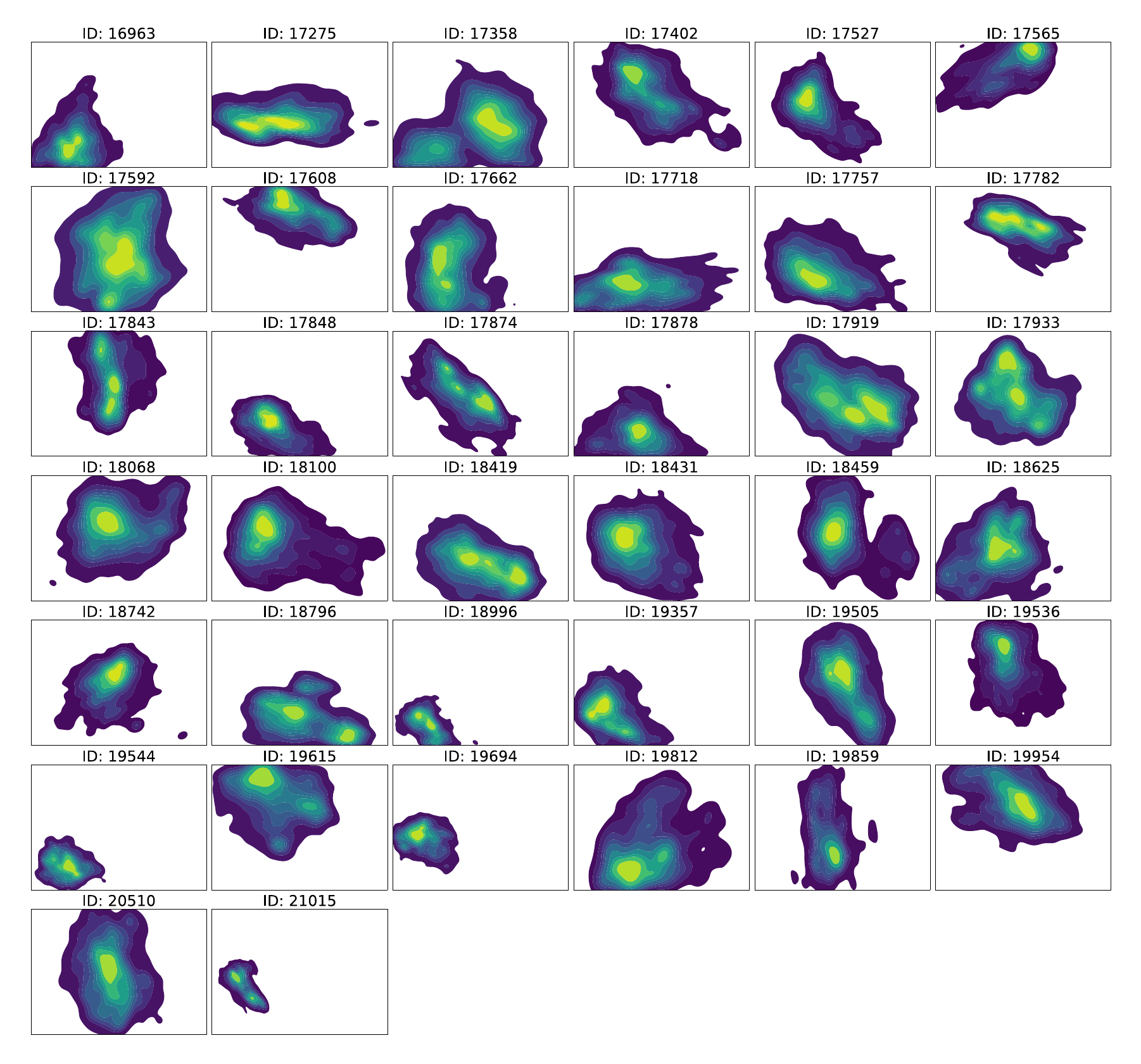}
  \caption{Gaze distribution of individual adults during visual exploration. Each subplot shows an adult's 2D gaze distribution (kernel density estimate) across a play session. Warmer colors indicate higher gaze density.}
  \label{fig:gaze_adults}
\end{figure}

In \cref{app:datasets}, we explain that the center of the frames is misaligned with respect to the stationary position of the eyes. To support this statement, \figureautorefname~\ref{fig:gaze_toddlers} and \figureautorefname~\ref{fig:gaze_adults} display the distribution of gaze locations for each toddler and adult, respectively. Brighter areas indicate higher frequencies of gaze fixation at those locations. The results indicate that the average gaze location is typically not centered with respect to the camera.

\subsection{Adults' central visual field experience also supports the learning of invariant object representations}
\label{appendix:adults}
\begin{figure}[ht]
  \centering
  \includegraphics[width=1\textwidth]{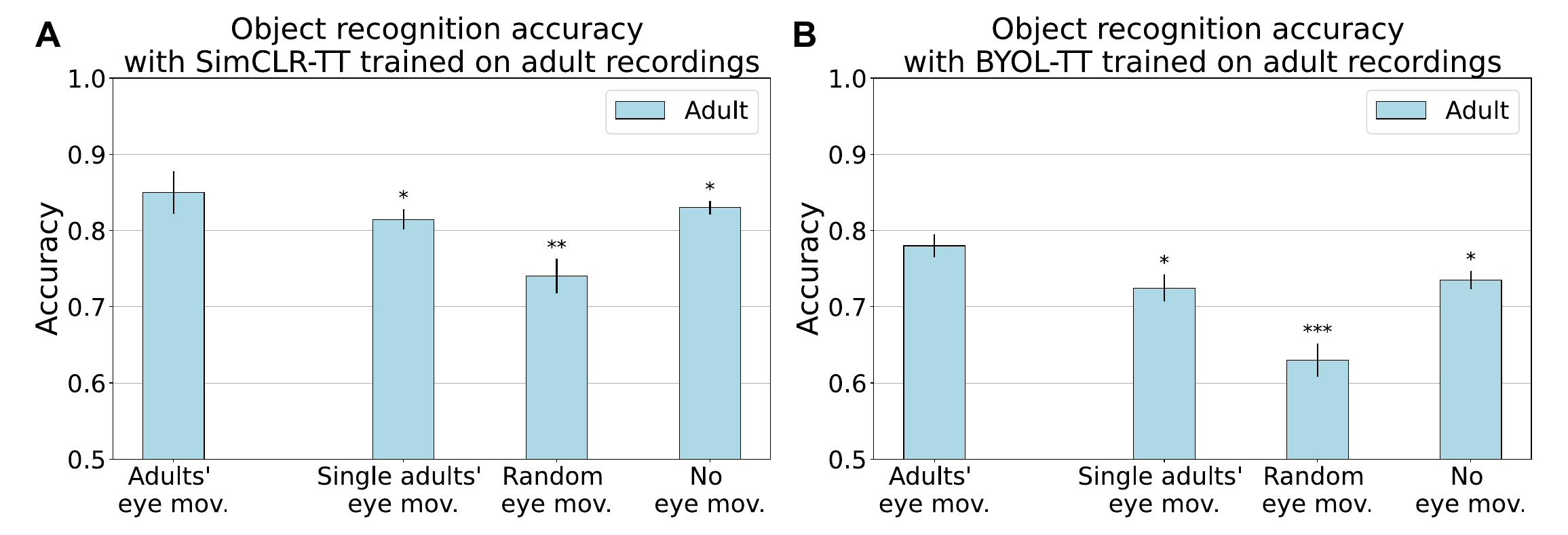}
  \caption{Linear object recognition accuracy of training on a single adult and all adults. {\bf (A)} and {\bf (B)} show the results for SimCLR-TT and BYOL-TT applied to the adult recordings, respectively. The error bars represent the standard deviation over three random seeds. When training on one adult, we show the average accuracy and the standard deviation over the 38 adults. We also provide the significance test results for adults' eye movements and other groups. ``***'' indicates $\text{p-value} < 0.001$, ``**'' indicates $\text{p-value} < 0.01$, and ``*'' indicates $\text{p-value} < 0.05$.}
  \label{fig:adult_trend}
\end{figure}

To examine the suitability of adults' looking behavior for learning viewpoint-invariant object representations, we replaced the toddlers' datasets with adults' data and applied the same evaluation procedure as described in \cref{sec:bio}. The results in \figureautorefname~\ref{fig:adult_trend} suggest that biologically inspired visual learning models like SimCLR-TT and BYOL-TT can also leverage adult gaze behavior to learn invariant object representations. The central visual field experience of one adult leads, on average, to representations almost as good as that of the combined central visual field experiences of all adults. These findings are in line with the conclusion about toddlers in \cref{sec:bio}.

\subsection{Impact of changing the self-supervised learning encoder}
\label{appendix:encoder}

\begin{figure}
  \centering
  \includegraphics[width=1\textwidth]{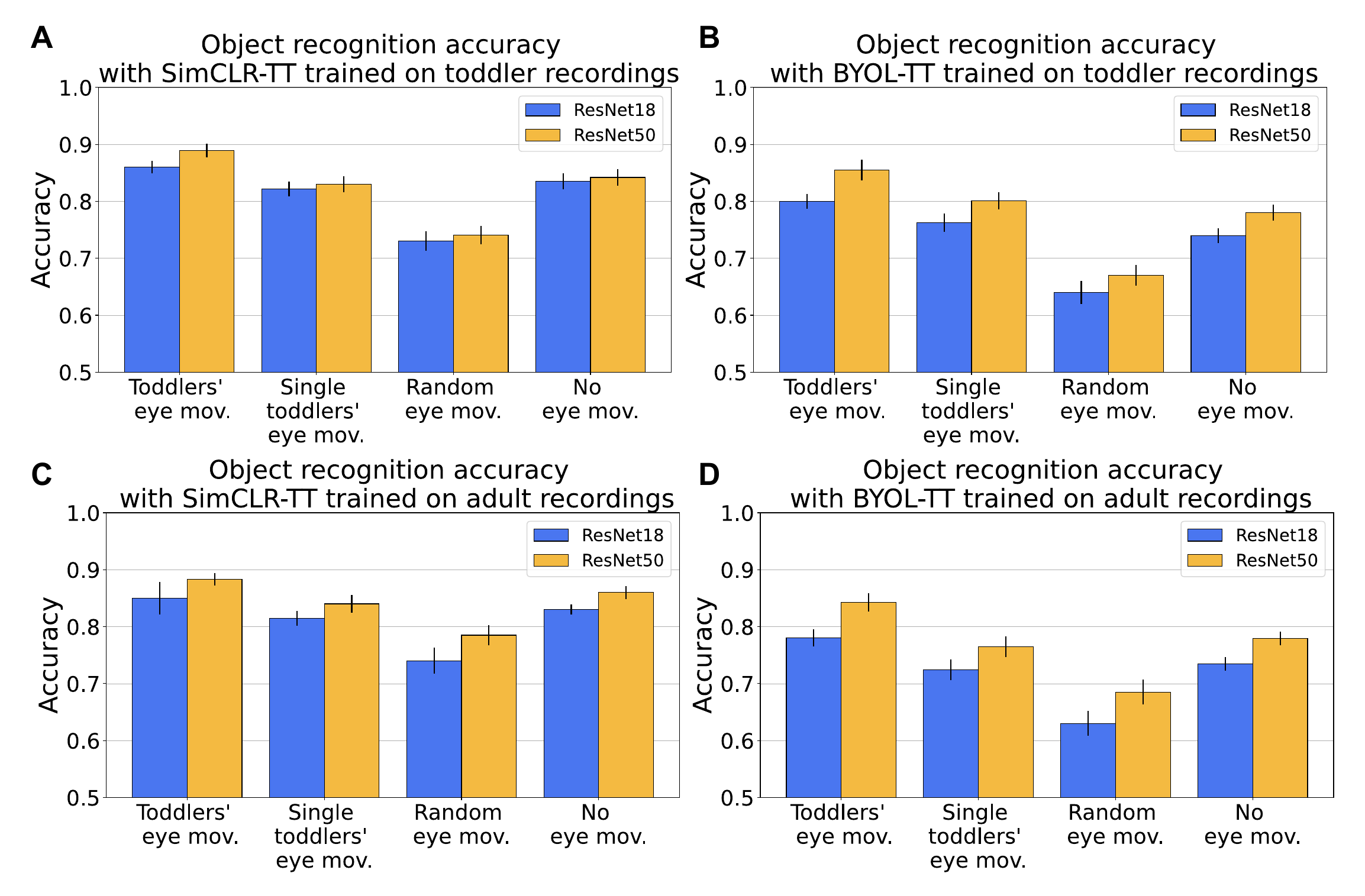}
  \caption{Different encoder training on individual toddler/adult and all toddlers/adults. {\bf (A)} and {\bf (B)} show the results for SimCLR-TT and BYOL-TT applied to toddlers' datasets, respectively; {\bf (C)} and {\bf (D)} present the corresponding results when switched to adults' datasets. Error bars represent one standard deviation over three random seeds.}
  \label{fig:byolencoder}
\end{figure}


We compared the accuracy of BYOL-TT and SimCLR-TT for ResNet18 vs.\ ResNet50 encoders on both the Toddlers' and Adults' datasets. As shown in \figureautorefname~\ref{fig:byolencoder}, the more complex ResNet50 encoder resulted in an improved accuracy, especially for BYOL-TT.

\subsection{Analysis of the class imbalance in the Objects Fixation dataset}
\label{appendix:imbalanced}

\begin{figure}[ht]
  \centering
  \includegraphics[width=1\textwidth]{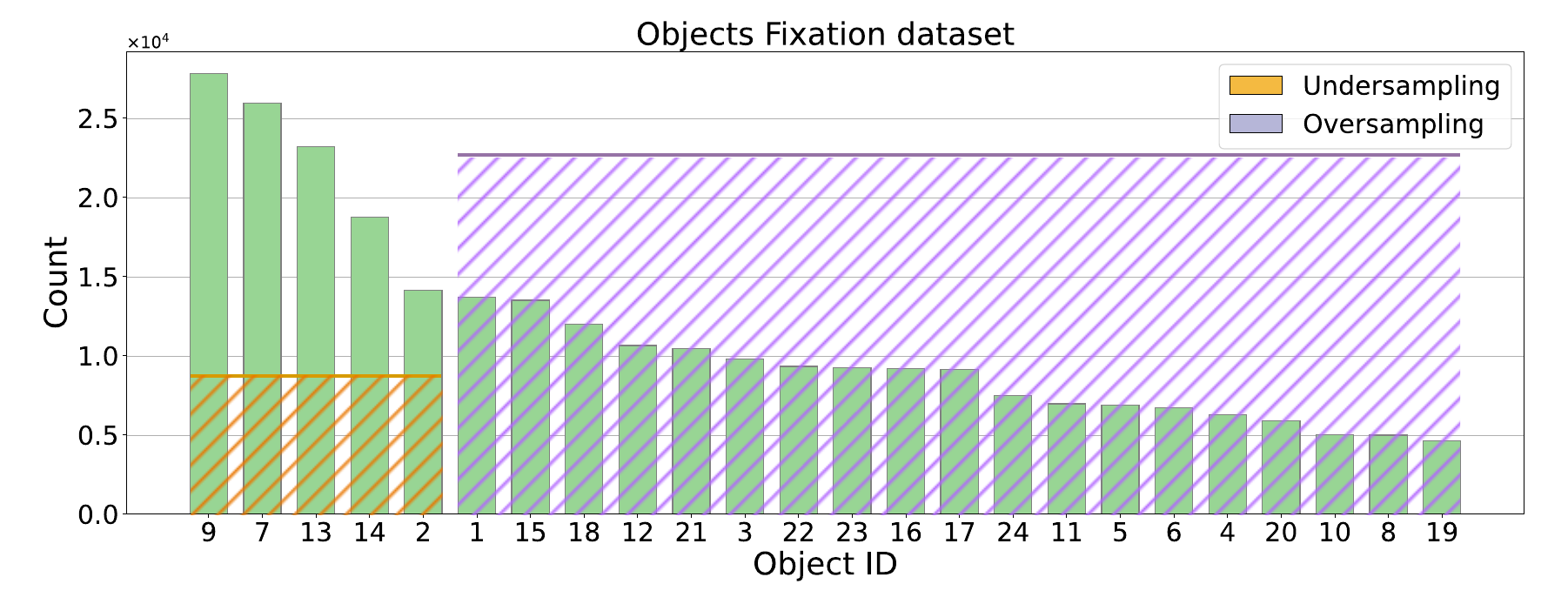}
  \caption{Sampling on the Objects Fixation dataset. Refer to the \tableautorefname~\ref{tab:24obj} and \tableautorefname~\ref{tab:24obj2} for the full object names corresponding to each ID. In the Top-5 classes, the yellow regions indicate the number of samples after undersampling. These, combined with all samples from the remaining classes, form the Undersampling dataset. The purple regions represent the number of samples after oversampling for the corresponding classes, which, together with all samples from the Top-5 classes, constitute the Oversampling dataset. The upper bound of the yellow regions corresponds to the average number of samples among the non-Top-5 classes, while the upper bound of the purple regions represents the average number of samples among the Top-5 classes. After applying undersampling and oversampling, the sizes of the three datasets are 204,462 (undersampling), 271,754 (original), and 527,449 (oversampling).}
  \label{fig:sampling}
\end{figure}

Toddlers and adults may spend different times looking at different objects. In \figureautorefname~\ref{fig:sampling}, we show the number of images per toy, sorted in descending order. We observe in the Objects Fixation dataset that some objects are more present than others.


To investigate the impact of this class imbalance, we adjusted the distribution of the Objects Fixation dataset while keeping the encoders pre-trained in \cref{sec:bio}. The linear classifier was then trained and tested on the adjusted datasets. We compared the results of two types of strategies that compensate for the class imbalance:

\textbf{Undersampling.} We applied random undersampling to reduce the number of samples in the top 5 categories, making their quantities similar to those of the other categories. 

\textbf{Oversampling.} Similarly, we applied random oversampling to increase the number of samples in the underrepresented categories to match the quantity of the top 5 classes. Compared to ``Undersampling'', this method results in duplicate samples in the dataset.

\begin{figure}[ht]
  \centering
  \includegraphics[width=1\textwidth]{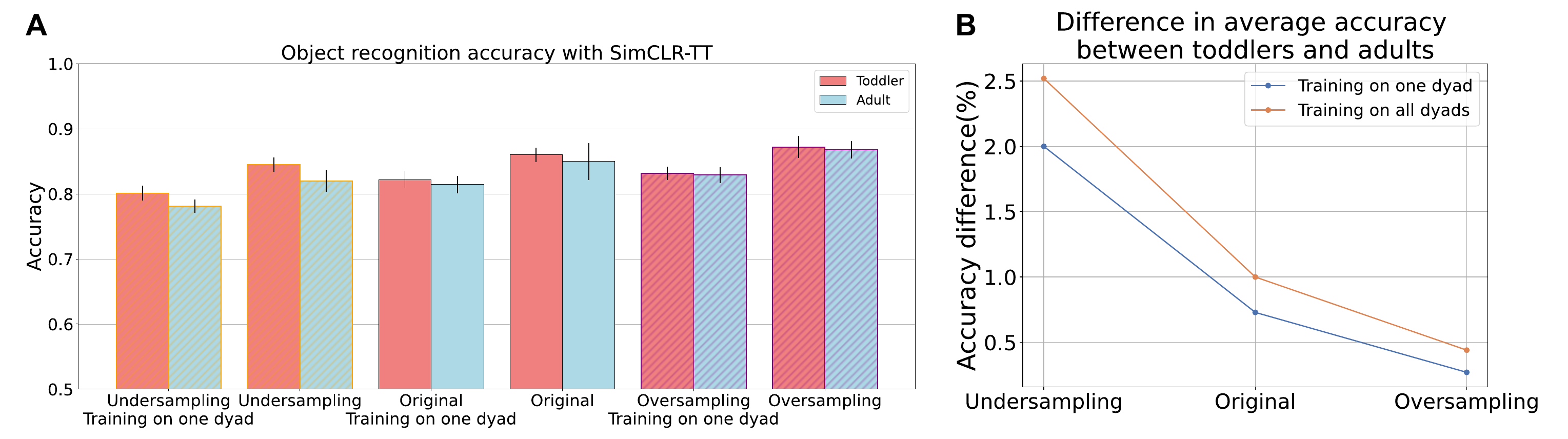}
\caption{Linear object recognition accuracy and the difference in accuracy between undersampling and oversampling. {\bf A.} We compared the recognition accuracy under different sampling methods. Additionally, we provide the test results after training on one dyad versus all dyads. Error bars represent the standard deviation over three random seeds; {\bf B.} The difference in recognition accuracy between toddlers and adults under different sampling methods. Here, we also compare the accuracy differences of the model trained on one dyad versus all dyads.} 
  \label{fig:samplingdiff}
\end{figure}

In \figureautorefname~\ref{fig:samplingdiff}, we observe that when the total sample size is reduced, the recognition accuracy of the models trained on Toddlers' and Adults' Eye movements dataset decreases, but the difference in their accuracy continues to widen. However, with more complex or balanced training, the model's generalization capacity improves, and the performance across toddlers and adults tends to converge, reducing the impact of differences in visual behaviors.


\begin{figure}[ht]
  \centering
  \includegraphics[width=0.94\textwidth]{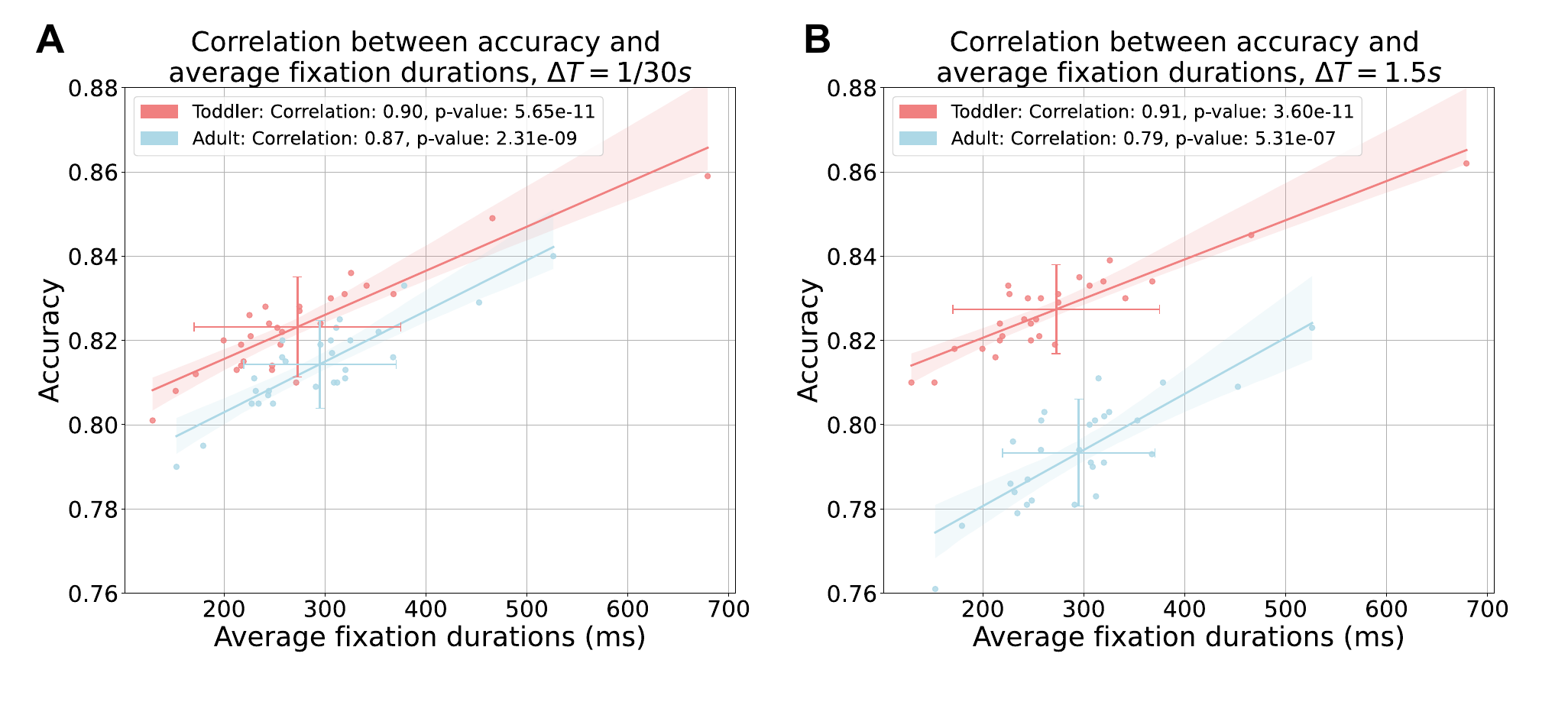}
  \caption{ $\Delta{T}$ highlights the differences between toddlers and adults. We observe variations in the test accuracy of models trained on the Toddlers' and Adults' Eye movements dataset under {\bf (A)} $\Delta{T}=\frac{1}{30}$s and  {\bf (B)} $\Delta{T}=1.5$s. We attached fitted regression lines, and the shaded areas show the 95\% confidence interval. The crosshairs represent the mean and standard deviation of the data values over the two axes.}
  \label{fig:enhancediff}
\end{figure}

\subsection{Further differences between toddlers and adults}
\label{appendix:difftod}
We provide additional evidence highlighting the differences between toddlers and adults. In \figureautorefname~\ref{fig:enhancediff}A and \figureautorefname~\ref{fig:enhancediff}B, we compare the changes in recognition accuracy for both toddlers and adults under different $\Delta{T}$ values. From the regression lines, the increasing $\Delta{T}$ amplifies the difference in recognition accuracy between training on Toddlers' and Adults' Eye movements datasets, consistent with the findings in \cref{sec:mod}. 


\begin{figure}[ht]
  \centering
  \includegraphics[width=1\textwidth]{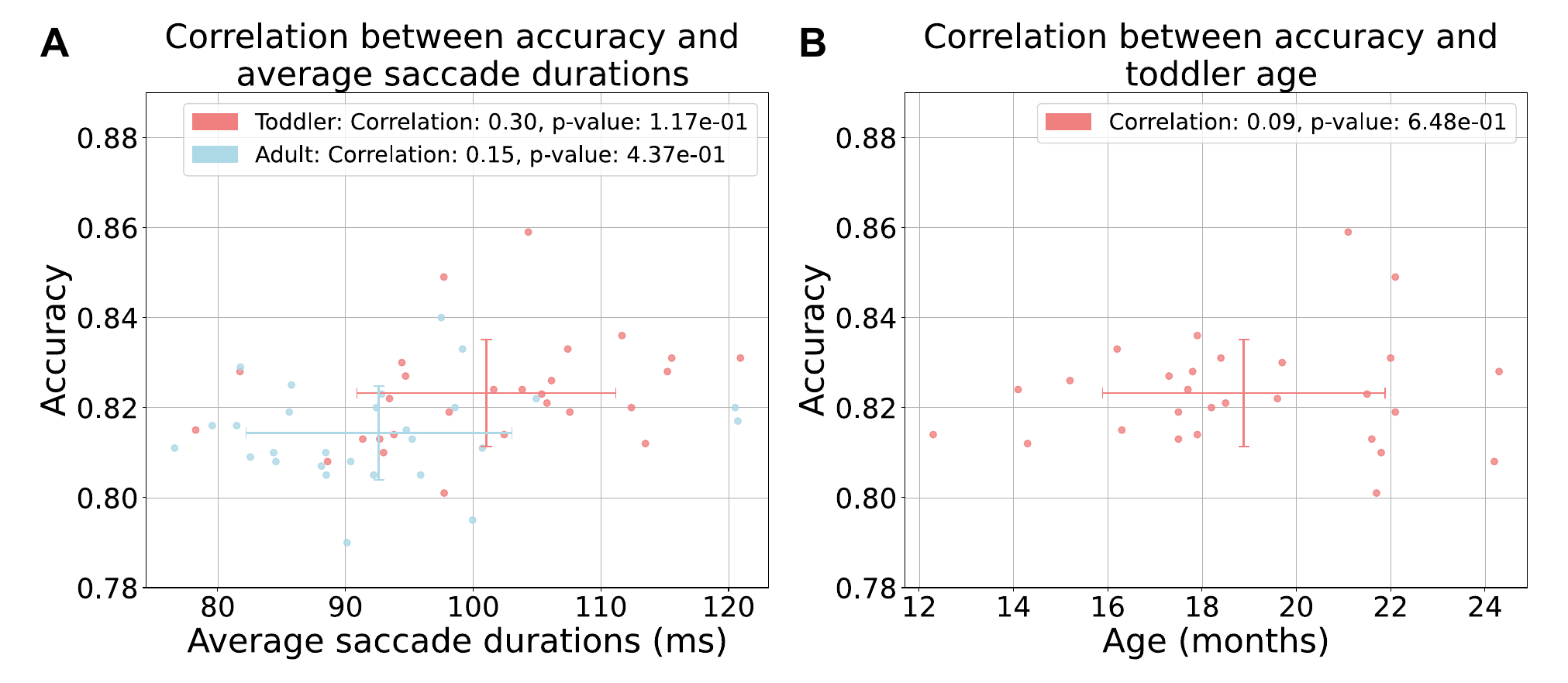}
  \caption{The impact of {\bf (A)} average saccade duration and {\bf (B)} toddler's age on recognition accuracy. The crosshairs represent the mean and standard deviation of the data values over the two axes.}
  \label{fig:saccade_age}
\end{figure}

\subsection{Study of saccade duration and age}
Here, we complement \cref{sec:metric} and study two additional metrics that may impact the performance of individual adults and toddlers, namely the average saccade duration and toddlers' age. According to \figureautorefname~\ref{fig:saccade_age}, we observe no significant correlation between the recognition accuracy and either the average saccade duration or toddlers' age. However, the youngest toddlers in the study were older than one year, and we can not rule out that younger babies may show different results.

\begin{figure}[ht]
  \centering
  \includegraphics[width=1\textwidth]{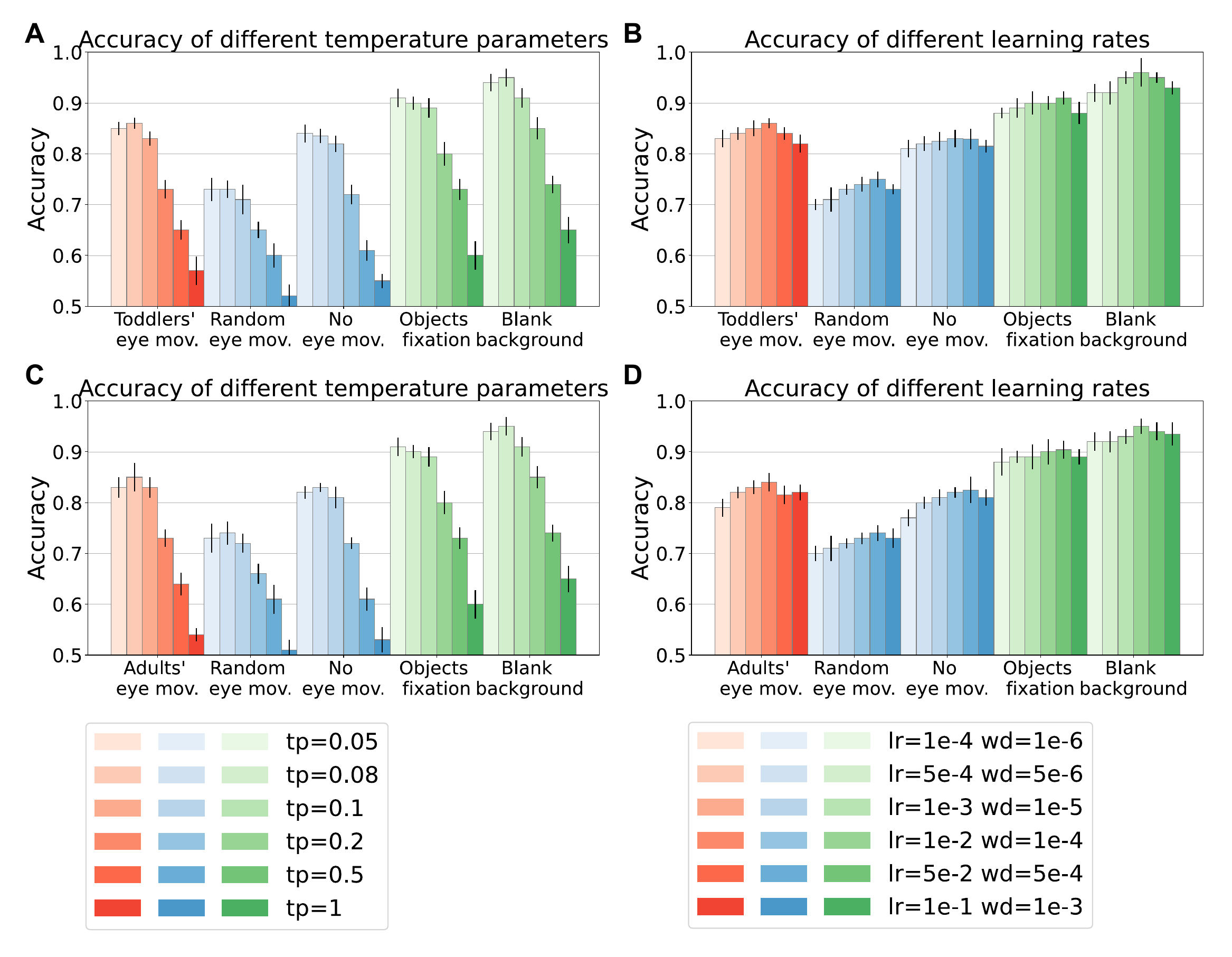}
  \caption{Object recognition accuracy across different hyper-parameter settings. {\bf (A)} and {\bf (B)} show the results for different learning rates and temperature parameters applied to toddlers' datasets, respectively, {\bf (C)} and {\bf (D)} present the corresponding results when switched to adults' datasets. The vertical bars represent the standard deviation over three random seeds.}
  \label{fig:hp}
\end{figure}

\subsection{Robustness under varying hyper-parameters}
\label{appendix:hp}
The learning rate (lr), weight decay (wd), and temperature (tp) used in our main experiments were selected as the best settings after a hyper-parameter search. To assess the robustness of our method, we conduct additional experiments where we set $\text{lr}=10^{-2}$, $\text{wd}=10^{-4}$, $\text{tp}=0.08$, and vary hyper-parameters individually. As shown in \figureautorefname~\ref{fig:hp}, changes in these hyper-parameters do not affect the conclusions presented in \cref{sec:bio}.

\section{Details of all toys and toddlers Data}
\label{appendix:data}
We provide information for all toys and toddlers participating in the study in \tableautorefname~\ref{tab:24obj}, \tableautorefname~\ref{tab:24obj2}, and \tableautorefname~\ref{tab:toddler}. The toddler ID represents an anonymized identifier for each toddler.  
\begin{table}[ht]
\caption{24 toys were used for toddler interaction. See \tableautorefname~\ref{tab:24obj2} for the second part. Among them, ``Library'' refers to those toys that were successfully recognized when the toddler calls any word from the corresponding row. However, these columns are not within the scope of the current study's discussion. The main focus is on the colors, shapes, or textures of these 24 toys, which are more likely to help toddlers differentiate between them. }
\vspace{0.5em}
\centering
\renewcommand\arraystretch{0.8}
\tabcolsep=0.15cm 
    \begin{tabular}{c|c|c|c}
        \toprule
        \textbf{Gazetag Naming} & \textbf{ICONS} & \textbf{ID} & \textbf{Library} \\ \hline
        \raisebox{6pt}{helmet} & \includegraphics[width=0.8cm]{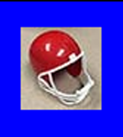} & \raisebox{6pt}{1} & \raisebox{6pt}{helmet, hat} \\
        \raisebox{6pt}{house} &  \includegraphics[width=0.8cm]{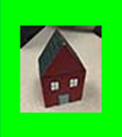} & \raisebox{6pt}{2} & \raisebox{6pt}{house, home} \\
        \raisebox{6pt}{bluecar} & \includegraphics[width=0.8cm]{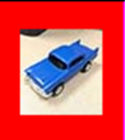} & \raisebox{6pt}{3} & \raisebox{6pt}{car} \\
        \raisebox{6pt}{rose} & \includegraphics[width=0.8cm]{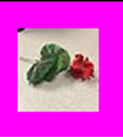} & \raisebox{6pt}{4} & \raisebox{6pt}{rose, flower, plant} \\
        \raisebox{6pt}{elephant} & \includegraphics[width=0.8cm]{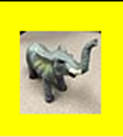} & \raisebox{6pt}{5} & \raisebox{6pt}{elephant} \\
        \raisebox{6pt}{snowman} & \includegraphics[width=0.8cm]{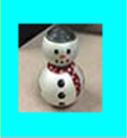} & \raisebox{6pt}{6} & \raisebox{6pt}{snowman} \\
        \raisebox{6pt}{rabbit} & \includegraphics[width=0.8cm]{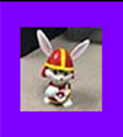} & \raisebox{6pt}{7} & \raisebox{6pt}{rabbit, bunny} \\
        \raisebox{6pt}{spongebob} & \includegraphics[width=0.8cm]{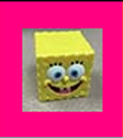} & \raisebox{6pt}{8} & \raisebox{6pt}{spongebob, block} \\
        \raisebox{6pt}{turtle} & \includegraphics[width=0.8cm]{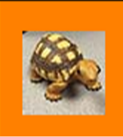} & \raisebox{6pt}{9} & \raisebox{6pt}{turtle, tortoise} \\
        \raisebox{6pt}{hammer} & \includegraphics[width=0.8cm]{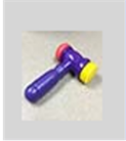} & \raisebox{6pt}{10} & \raisebox{6pt}{hammer, tool, mallet} \\
        \raisebox{6pt}{ladybug} & \includegraphics[width=0.8cm]{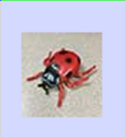} & \raisebox{6pt}{11} & \raisebox{6pt}{bug, insect, ladybug, beetle} \\
        \raisebox{6pt}{mantis} & \includegraphics[width=0.8cm]{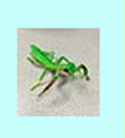} & \raisebox{6pt}{12} & \raisebox{6pt}{bug, insect, praying mantis, mantis, grasshopper} \\
        \bottomrule
    \end{tabular}
\label{tab:24obj}
\end{table}

\begin{table}[]
\caption{24 toys were used for toddler interaction. See \tableautorefname~\ref{tab:24obj} for the first part.}
\vspace{0.5em}
\centering
\renewcommand\arraystretch{0.8}
\tabcolsep=0.15cm 
    \begin{tabular}{c|c|c|c}
        \toprule
        \textbf{Gazetag Naming} & \textbf{ICONS} & \textbf{ID} & \textbf{Library} \\ \hline
         \raisebox{6pt}{greencar} & \includegraphics[width=0.8cm]{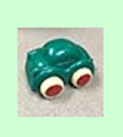} & \raisebox{6pt}{13} & \raisebox{6pt}{car} \\
        \raisebox{6pt}{saw} & \includegraphics[width=0.8cm]{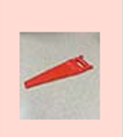} & \raisebox{6pt}{14} & \raisebox{6pt}{saw, tool} \\
        \raisebox{6pt}{doll} & \includegraphics[width=0.8cm]{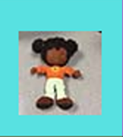} & \raisebox{6pt}{15} & \raisebox{6pt}{baby, baby doll, girl, doll} \\
        \raisebox{6pt}{phone} & \includegraphics[width=0.8cm]{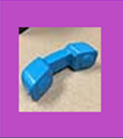} & \raisebox{6pt}{16} & \raisebox{6pt}{phone, telephone} \\
        \raisebox{6pt}{rubiks} & \includegraphics[width=0.8cm]{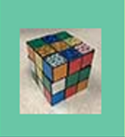} & \raisebox{6pt}{17} & \raisebox{6pt}{block, rubiks cube, rubiks, cube} \\
        \raisebox{6pt}{shovel} & \includegraphics[width=0.8cm]{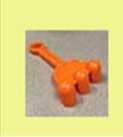} & \raisebox{6pt}{18} & \raisebox{6pt}{rake, shovel, tool} \\
        \raisebox{6pt}{bigwheels} & \includegraphics[width=0.8cm]{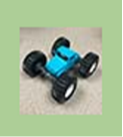} & \raisebox{6pt}{19} & \raisebox{6pt}{truck, jeep, bigwheel, car} \\
        \raisebox{6pt}{whitecar} & \includegraphics[width=0.8cm]{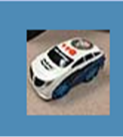} & \raisebox{6pt}{20} & \raisebox{6pt}{car, policecar} \\
        \raisebox{6pt}{ladybugstick} & \includegraphics[width=0.8cm]{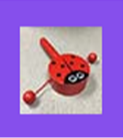} & \raisebox{6pt}{21} & \raisebox{6pt}{ladybug, bug} \\
        \raisebox{6pt}{purpleblock} & \includegraphics[width=0.8cm]{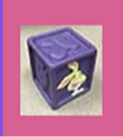} & \raisebox{6pt}{22} & \raisebox{6pt}{block, cube} \\
        \raisebox{6pt}{bed} & \includegraphics[width=0.8cm]{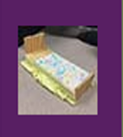} & \raisebox{6pt}{23} & \raisebox{6pt}{bed} \\
        \raisebox{6pt}{clearblock} & \includegraphics[width=0.8cm]{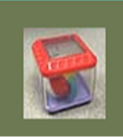} & \raisebox{6pt}{24} & \raisebox{6pt}{block, cube} \\
        \bottomrule
    \end{tabular}
\label{tab:24obj2}
\end{table}

\begin{table}[]
\caption{Information for each participating toddler (Anonymized). IDs marked with `` * '' indicate inclusion in \cref{sec:metric}. Frame Count denotes the total video frames used; Video Length refers to the recording duration. Age represents the participant's age at recording. Gender is marked as M (male) or F (female). Resolution indicates the video resolution from the head-mounted camera.}
\vspace{0.5em}
\tabcolsep=0.2cm
\centering
\renewcommand\arraystretch{1.3}
\begin{tabular}{c|c|c|c|c|c}
\toprule
\raisebox{0.1cm}{\textbf{Toddler ID}} & \raisebox{0.1cm}{\textbf{Frame Count}} & \textbf{\shortstack{Video Length \\ (min. : sec.)}} & \textbf{\shortstack{Age \\ (months)}}   & \raisebox{0.1cm}{\textbf{Gender}} & \textbf{\shortstack{Resolution \\ (px$^2$)}} \\ \hline
16963     & 16440       & 9:07         & 20.7  & M      & $720 \times 480$    \\
17275     & 9120        & 5:04         & 18.2  & F      & $720 \times 480$     \\
17358     & 18930       & 10:31        & 18.8  & M      & $720 \times 480$     \\
17402     & 27636       & 15:21        & 19.2  & M      & $640 \times 480$    \\
17527*     & 15242       & 8:28         & 21.5  & M      & $640 \times 480$    \\
17565*     & 14864       & 8:15         & 19.7  & F      & $640 \times 480$    \\
17592*     & 16116       & 8:58         & 18.2  & M      & $640 \times 480$    \\
17608*     & 18059       & 10:02        & 21.8  & F      & $640 \times 480$    \\
17662*     & 14553       & 8:05         & 15.2  & F      & $640 \times 480$    \\
17718     & 11850       & 6:35         & 18.1  & F      & $720 \times 480$     \\
17757*     & 19661       & 10:55        & 21.7  & F      & $640 \times 480$    \\
17782*     & 9035        & 5:01         & 22.1  & F      & $640 \times 480$    \\
17843*     & 18209       & 10:07        & 19.6  & F      & $640 \times 480$    \\
17848*     & 21111       & 11:43        & 18.4  & F      & $640 \times 480$    \\
17874*     & 17429       & 9:41         & 17.8  & M      & $640 \times 480$    \\
17878*     & 20018       & 11:08        & 17.5  & F      & $640 \times 480$    \\
17919*     & 18596       & 10:20        & 22.1  & M      & $640 \times 480$    \\
17933*     & 14457       & 8:02         & 17.9  & F      & $640 \times 480$    \\
18068*     & 7976        & 4:26         & 17.9  & M      & $640 \times 480$    \\
18100*     & 14982       & 8:19         & 16.3  & F      & $640 \times 480$    \\
18419*     & 28253       & 15:41        & 17.3  & M      & $640 \times 480$    \\
18431*     & 11575       & 6:26         & 22    & M      & $640 \times 480$    \\
18459*     & 7231        & 4:01         & 16.2  & F      & $640 \times 480$    \\
18625*     & 18209       & 10:07        & 24.3  & F      & $640 \times 480$    \\
18742*     & 19018       & 10:34        & 17.7  & M      & $640 \times 480$    \\
18796*     & 11672       & 6:30         & 24.2  & M      & $640 \times 480$    \\
18996     & 12466       & 6:56         & 15.9  & F      & $320 \times 240$    \\
19357*     & 8834        & 4:54         & 17.5  & M      & $640 \times 480$    \\
19505*     & 18397       & 10:13        & 18.5  & M      & $640 \times 480$    \\
19536*     & 18370       & 10:13        & 21.1  & M      & $640 \times 480$    \\
19544     & 9151        & 5:05         & 13.8  & F      & $320 \times 240$     \\
19615*     & 13351       & 7:25         & 14.1  & M      & $640 \times 480$    \\
19694     & 10801       & 6:00         & 15.2  & M      & $320 \times 240$     \\
19812*     & 9918        & 5:31         & 21.6  & M      & $640 \times 480$    \\
19859     & 7360        & 4:05         & 14.4  & M      & $640 \times 480$    \\
19954*     & 9201        & 5:07         & 12.3  & F      & $640 \times 480$    \\
20510*     & 11865       & 6:35         & 14.35 & M      & $640 \times 480$    \\
21015     & 9566        & 5:19         & 13    & M      & $320 \times 240$     \\ \bottomrule
\end{tabular}
\label{tab:toddler}
\end{table}




\end{appendices}



\end{document}